**RESEARCH ARTICLE**

**OPEN ∂ ACCESS**

# Planning of efficient trajectories in robotized assembly of aerostructures exploiting kinematic redundancy

Federica Storiale[*], Enrico Ferrentino[**], and Pasquale Chiacchio[***]

Department of Computer Engineering, Electrical Engineering and Applied Mathematics (DIEM), University of Salerno, Fisciano, SA 84084, Italy

**Abstract.** Aerospace production volumes have increased over time and robotic solutions have been progressively introduced in the aeronautic assembly lines to achieve high-quality standards, high production rates, flexibility and cost reduction. Robotic workcells are sometimes characterized by robots mounted on slides to increase the robot workspace. The slide introduces an additional degree of freedom, making the system kinematically redundant, but this feature is rarely used to enhance performances. The paper proposes a new concept in trajectory planning, that exploits the redundancy to satisfy additional requirements. A dynamic programming technique is adopted, which computes optimized trajectories, minimizing or maximizing the performance indices of interest. The use case is defined on the LABOR (Lean robotized AssemBly and cOntrol of composite aeRostructures) project which adopts two cooperating six-axis robots mounted on linear axes to perform assembly operations on fuselage panels. Considering the needs of this workcell, unnecessary robot movements are minimized to increase safety, the mechanical stiffness is maximized to increase stability during the drilling operations, collisions are avoided, while joint limits and the available planning time are respected. Experiments are performed in a simulation environment, where the optimal trajectories are executed, highlighting the resulting performances and improvements with respect to non-optimized solutions.

## 1 Introduction

Nowadays, the huge volumes in manufacturing industries have brought to an increment in the employment of autonomous systems performing the hardest and repeatable operations, in order to increase the overall efficiency of the production lines. As a consequence, robotized solutions are frequently adopted to obtain a higher level of automation, while guaranteeing high quality results. On the one hand, the aerospace sector, as highlighted in [1], is the least automated because of the large and complex systems to handle and the wide variety of activities to be carried out during the production phases, including drilling, sealing, fastening, inspection, coating, painting and material handling [2]. On the other hand, even in this context, in agreement with the global trend, production volumes have increased in the last years, requiring automatic solutions where robots perform such operations.

Typically, robotized solutions in large industrial plants have the common characteristic of employing robots mounted on slides, i.e. linear axes, to increase their workspace and allow them to cover wide areas. These linear axes introduce additional degrees of freedom that yield kinematic redundancy, i.e. there is an infinite number of joint configurations corresponding to the same pose of the end-effector.

Kinematic redundancy is also the key in applications involving mobile manipulators [3], that are versatile systems, capable of performing different kinds of tasks. Their flexibility comes with an increased complexity due, among other things, to the capability of handling the kinematic redundancy efficiently. This characteristic has been the main impediment to the spreading of mobile robots in the aerospace manufacturing lines. In fact, in order to simplify the setup of the workcell, the existing solutions foresee to neglect the extra degrees of freedom during trajectory planning, using them exclusively to increase the workspace of the robots.

Existing robotized solutions in aerostructures manufacturing are typically complex, heavy, rigid and expensive, since high-payload robots are adopted and the demanding requirements of the assembly operations yield an increase of cost. Especially in the case of regional aircraft industry, assembly of fuselage aerostructures is largely a manual process because of the need for high accuracy, which is not achievable by the common industrial robots. In fact, this is typically guaranteed through external metrology systems that, however, increase the overall cost of the system and reduce its flexibility.

* e-mail: fstoriale@unisa.it
** e-mail: eferrentino@unisa.it
*** e-mail: pchiacchio@unisa.it

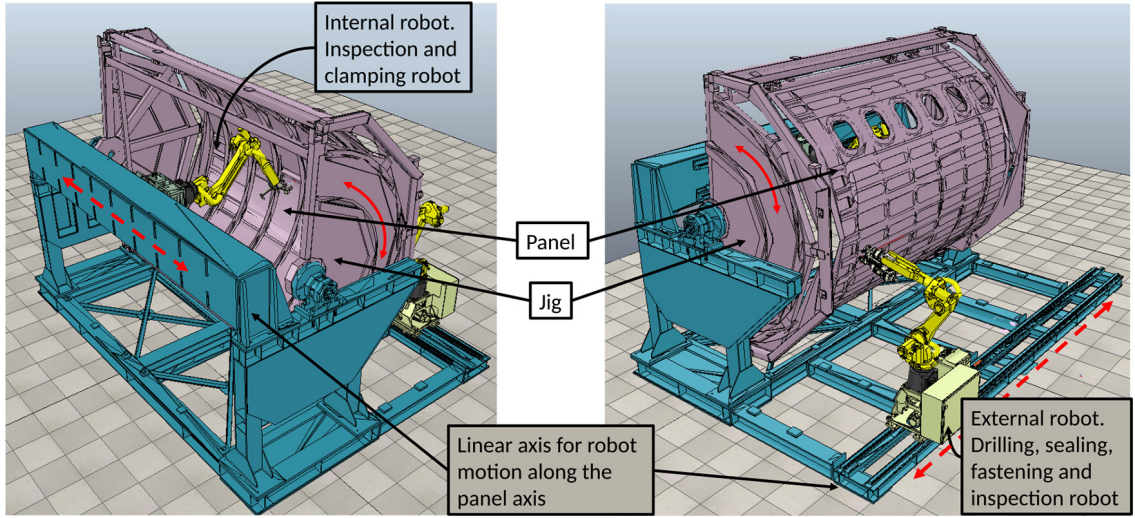

**Fig. 1.** LABOR cell represented in the CoppeliaSim simulator.

Furthermore, a complete assembly cycle requires to perform the mentioned tasks on thousands of holes per aircraft, thus tight time constraints are usually imposed on each single operation to keep the production rates high. Also, since in most cases the human intervention is still required, the automated workcell must be adapted for human cooperation [1,4,5]. In particular, a higher degree of safety is necessary, avoiding dangerous robot movements and configurations.

This paper concerns the employment of small and medium sized high-precision robots, whose workspace is augmented through the introduction of extra degrees of freedom. They guarantee higher flexibility and speed, naturally contain costs and can be programmed to increase production quality and safety. Kinematic redundancy is exploited to optimize one or more performance indices of interest for aerostructures assembly, as well as to respect all the typical constraints that characterize such applications.

Finally, as highlighted in [6], industrial robots are often programmed by using the robot-specific teach pendant, which is a very slow and not efficient solution and is not suitable especially for mobile robots moving in a dynamic environment. Handling redundancy allows to define tasks directly in the workspace, pushing the programming interface to a higher level of abstraction, leading to greater flexibility and efficiency in the workcell management.

### 1.1 Motivation of the paper

The case study of this paper comes from the European project LABOR (Lean robotized AssemBly and cOntrol of composite aeRostructures) [7]. Its objective is to build an automated solution for the assembly process of fuselage parts, such as skins, stringers, frames and door surround components, where drilling, sealing, fastening and inspection are performed by small/medium size robots, in replacement of the large and heavy robots that characterize the state of art in aeronautical machining cells.

A 3D reconstruction of the LABOR cell is provided in Figure 1. The panel is mounted on a rotating jig, that holds and orients the panel during the assembly operations performed by two cooperating six-axis robots mounted on the two sides of the panel, namely: the external and internal (with respect to the curvature of the panel) robot. The former performs drilling, sealing, fastening and inspection from the hole entry side while the latter performs the inspection from the hole exit side and applies a clamping force during the drilling phase. Each robot is mounted on a platform moving on a linear axis (the 7-th axis), allowing for the movement along the length of the panel. The combined movement of the robots, the slide and the jig allows to cope with the size of the panel which is much larger than the workspace of the two six-axis robots.

The nominal working sequence starts with the internal robot inspecting the area, performing the referencing and computing the drilling coordinates which are sent to the external robot. Then, the external robot starts drilling while the internal robot applies the counterthrust force. At the end of the drilling operations, the external robot inspects the drilled hole. Such operations are repeated for each hole of the sequence. Once all the holes have been drilled, the external robot mounts the sealing and fastening tool and goes back to the first hole of the sequence to start sealing and fastening. At the end, if it is required, the internal robot performs the inspection of the fastened holes. The work to be performed on every hole must not exceed 30 s (excluding after-fastening inspection) in order to respect the overall cycle time for the whole panel.

The current setup foresees that the trajectories tracked by the robot during the assembly operations (e.g. from one hole to the other) are directly assigned in the joint space, by keeping the jig and slide fixed. In particular, jig-slide positions are found corresponding to best working areas for the robots to improve the quality of the assembly operations. These are found through some heuristics such as placing the slide right in front of the hole with the robot tool close to the base and the arm less elongated. This way, the redundancy introduced by the presence of the slide is not exploited, resulting in several shortcomings.

In the current setup, holes are in vertical sequences, such that the slide is not moved during the operations on a single sequence. This limits the performances, preventing to efficiently operate on different hole patterns which are needed for some specific panel areas, e.g. doors, windows. Since large and geometrically complex tools are mounted at the robot end-effector, collisions with the panel and with the robot itself can easily occur. Hence the system must be equipped with collision avoidance algorithms, which are even more important when human operators access the cell during the working phase. In addition, the interaction with the panel might not be stable, since slippage and mechanical deviations may occur. Also, the time constraints imposed by the process have to be respected, as well as joint limits imposed by the robot manufacturer.

Our goal is to deal with all these aspects by planning trajectories that allow for more complex hole pattern geometries, avoid unnecessary robot movements through minimum joint displacements, prevent dangerous robot configurations through collisions checking, increase stability, by stiffness maximization, during drilling, while respecting joint limits and being compliant with the available planning time. The joint space configurations are computed by simply defining the task in the workspace (e.g. the position and orientation coordinates of the hole to be drilled) and then optimizing, in the allotted time and in a global manner, the postures that the robot has to assume to make the end-effector reach the assigned positions.

Trajectories are optimized using the approach proposed in [8] which is based on a dynamic programming (DP) algorithm for planning robot trajectories in the joint space, exploiting kinematic redundancy, in order to satisfy additional requirements and effectively increase the efficiency and the flexibility of the whole system. The slide will not be kept fixed, but its motion will be planned as part of the optimization process, treating it as an additional joint to achieve more efficient configurations.

In Section 2, we recall the notion of kinematic redundancy and present the dynamic programming approach, analyzing the technique of the force ellipsoids for the stiffness maximization. Then, in Section 3, experimental results are provided, comparing them with the traditional approach. At the end, in Section 4, conclusions and possible future developments are discussed.

## 2 Problem formulation

### 2.1 Redundancy resolution

A manipulator is defined as *kinematically redundant* when the number $m$ of task constraints is lower than the number $n$ of degrees of freedom provided by the manipulator's kinematic chain. $r = n - m$ is termed *degree of redundancy*.

Let $\mathbf{q} = [q_1 \ q_2 \ ... \ q_n]^T$ be the $n \times 1$ vector of joint positions representing the configuration of the manipulator and $\mathbf{x} = [\mathbf{p} \ \boldsymbol{\phi}]^T$ the $m \times 1$ vector of task position $\mathbf{p}$ and orientation $\boldsymbol{\phi}$ expressing the end-effector frame $\mathcal{T}_e$ coordinates with respect to the base frame $\mathcal{T}_b$. Considering a task described by six variables, the position is $\mathbf{p} \in \mathbb{R}^3$ and the orientation $\boldsymbol{\phi}(\mathbf{R}) \in \mathbb{R}^3$ is expressed through the set of Euler angles [9] extracted from the $3 \times 3$ rotation matrix $\mathbf{R} \in SO(3)$ from $\mathcal{T}_b$ to $\mathcal{T}_e$. The mapping from the joint space to the task space is performed through the non-linear vectorial function $\mathbf{k} : \mathbb{R}^n \to SE(3)$, where $SE(3) = \{(\mathbf{p}, \mathbf{R}) : \mathbf{p} \in \mathbb{R}^3, \mathbf{R} \in SO(3)\} = \mathbb{R}^3 \times SO(3) = \mathbb{R}^m$. The direct kinematic equation, representing the path constraint, is expressed as

$$\mathbf{x}(t) = \mathbf{k}(\mathbf{q}(t)), \qquad (1)$$

where $t \in [0, T]$ denotes the time and $T$ is the trajectory duration.

When the task is assigned in the task space, the kinematic equation (1) has to be inverted in order to find the joint positions allowing to fulfill such a task, that is

$$\mathbf{q}(t) = \mathbf{k}^{-1}(\mathbf{x}(t)). \qquad (2)$$

For a redundant robot, the inverse kinematics problem in (2) admits, in general, an infinite set of solutions, i.e. infinite joint positions that keep the end-effector motionless. This means that it is possible to optimize across such solutions to achieve other objectives, besides respecting the task constraint. This optimization process is usually referred to as *redundancy resolution*.

According to [10], the infinite set of solutions can be parametrized with $r$ functions of the joint positions. Let us call the vector of these functions $\mathbf{u}$ and add it to the direct kinematic equations $\mathbf{k}$, so as to obtain:

$$\begin{bmatrix} \mathbf{x}(t) \\ \mathbf{u}(t) \end{bmatrix} = \begin{bmatrix} \mathbf{k}(\mathbf{q}(t)) \\ \mathbf{k}_u(\mathbf{q}(t)) \end{bmatrix} = \mathbf{k}_a(\mathbf{q}(t)) \qquad (3)$$

where $\mathbf{k}_u : \mathbb{R}^n \to \mathbb{R}^r$ is the forward kinematics of some joint combinations and $\mathbf{k}_a$ is the *augmented kinematics*. By differentiating (3), we obtain:

$$\begin{bmatrix} \dot{\mathbf{x}}(t) \\ \dot{\mathbf{u}}(t) \end{bmatrix} = \begin{bmatrix} \mathbf{J}(\mathbf{q}(t)) \\ \mathbf{J}_u(\mathbf{q}(t)) \end{bmatrix} \dot{\mathbf{q}}(t) = \mathbf{J}_a(\mathbf{q}(t)) \dot{\mathbf{q}}(t) \qquad (4)$$

where $\mathbf{J} = \frac{\partial \mathbf{k}}{\partial \mathbf{q}}$ is the task Jacobian, $\mathbf{J}_u = \frac{\partial \mathbf{k}_u}{\partial \mathbf{q}}$ is the redundancy parameter Jacobian and $\mathbf{J}_a = \frac{\partial \mathbf{k}_a}{\partial \mathbf{q}}$ is the Jacobian matrix of the augmented kinematics. The problem (3) is squared and can be inverted out of singularities, when $\mathbf{u}$ is given. In the most simple case, $\mathbf{u}$ can be made of $r$ joint positions, whose selection, however, is not trivial. For certain tasks, the manipulator may still be redundant (and present infinite inverse kinematics solutions), even though $r$ joints are fixed. In such cases, the selected joints are not representative of the redundancy space, i.e. the null space of the Jacobian $\mathcal{N}(\mathbf{J})$, and $\mathbf{J}_a$ is rank-deficient for some trajectory point. Therefore,

we must ensure that

$$\text{rank}(\mathbf{J}_a) = n \quad (5)$$

for each trajectory point. This requires to choose the redundant joints such that

$$\mathcal{R}(\mathbf{J}^T) \cap \mathcal{R}(\mathbf{J}_u^T) = \emptyset \quad (6)$$

where $\mathcal{R}(\mathbf{M})$ represents the range space of a generic matrix $\mathbf{M}$ [10].

A solution that satisfies (6) is not easy to find analytically, especially for complex robots. Typically, in practice, the choice is made empirically, depending on the experience of the programmer. In our case, it can be verified a posteriori, numerically or through the graphical representation of the null space that the dynamic programming approach itself provides. If the joints are correctly selected, a finite number of solutions to the inverse kinematic problem is retrieved by inverting (3). In particular, the number of solutions depends on the robot type (i.e. planar, regional, spherical, spatial) and on the imposed constraints.

As will be clear in Section 2.2, the Jacobian is only used to verify the representativeness of the redundant joints, since the first order kinematics is not needed. Operating at joint position level is an important characteristic of the dynamic programming approach, resulting to be immune to singularities.

## 2.2 Redundancy optimization with dynamic programming

In [8], redundancy resolution is addressed through discrete dynamic programming. In this paper, we briefly recall this framework and extend it to the specific objectives and constraints that characterize the assembly of aerostructures in the LABOR project.

In the case of the LABOR cell, where the end-effector tools must have a specific orientation due to their shape and volume, the task variables are always six ($m = 6$), even for drilling, which is usually described by only five variables [11]. Given a six-axis robot mounted on an additional linear axis ($n = 7$), we thus have $r = 1$. In this paper, we provide a formulation for this particular case, but the reader may realize that it can be easily extended to cases where $r > 1$.

Let us consider the trajectory $\mathbf{x}(t)$, and discretize $t$ in its domain with $N_i + 1$ samples, with sampling interval $\tau = \frac{T}{N_i}$. Then, let us parametrize the redundancy by joint selection, so that $\mathbf{u} = u = q_i$, where $i$ is the $i$-th joint (the selected one). We discretize $u$ with $N_j + 1$ samples in its physical domain that depends on joint limits.

The augmented forward kinematics (3) can then be inverted in this discrete domain, for each single sample of $\mathbf{x}(t)$ and $u(t)$:

$$\mathbf{q}_{j,g}(i) = \mathbf{k}_a^{-1}(\mathbf{x}(i), u_j(i)) \quad (7)$$

where $i$ and $j$ are the indices that span the samples of the time and the redundancy parameter respectively, and $g = 1, ..., N_g$ is the index accounting for the presence of multiple inverse kinematic solutions when the redundancy parameter is given.

As mentioned in Section 1, the joint configurations and their derivatives must satisfy joint limits (position and velocity) and avoid self-collisions and collisions with the surrounding environment. Joint limits are formalized as follows:

$$\mathbf{q}_{\min} \leq \mathbf{q}(i) \leq \mathbf{q}_{\max} \quad (8)$$

$$\dot{\mathbf{q}}_{\min} \leq \dot{\mathbf{q}}(i) \leq \dot{\mathbf{q}}_{\max} \quad (9)$$

Collision constraints are checked referring to the geometric shapes [12] of each robot joint $S(q_j(i))$ and of the environment $S_e$, in such a way that the following relationships are always verified:

$$S(q_j(i)) \cap S(q_k(i)) = \emptyset \; \forall j, k = 1, ..., n \; \text{with} \; j \neq k \quad (10)$$

$$S(q_j(i)) \cap S_e = \emptyset \; \forall j = 1, ..., n \quad (11)$$

Thus, we can define the set $\mathcal{A}_i$ of admissible $\mathbf{q}$ at waypoint $i$ which takes the role of accepting only those configurations satisfying (8), (10) and (11). Similarly, the joint velocity limits $\dot{\mathbf{q}}$ can be accounted for with the set $\mathcal{B}_i(\mathbf{q}(i))$ which, in general, is time-dependent, as well as state dependent, i.e.

$$\mathcal{A}_i = \{\mathbf{q}(i) : (8), (10), (11) \text{ hold}\}$$

$$\mathcal{B}_i = \begin{cases} \dot{\mathbf{q}}(i) = \dfrac{\mathbf{q}(i) - \mathbf{q}(i-1)}{\tau} : \\ \mathbf{q}(i) \in \mathcal{A}_i, \mathbf{q}(i-1) \in \mathcal{A}_{i-1}, (9) \text{ holds} \end{cases} \quad (12)$$

where the backward Euler approximation has been used for discrete-time derivatives.

Once $\mathbf{q}_{j,g}(i)$ is available for each single value of $i, j$ and $g$, the objective is to find the optimal sequence of inputs that minimizes or maximizes a given cost function. Let us consider a generic cost function $I$ that is computed incrementally by summing up the local costs $l$ for each $i$:

$$I(N_i) = \psi(\mathbf{q}(0)) + \sum_{i=1}^{N_i} l(\mathbf{q}(i), \mathbf{q}(i-1)) \quad (13)$$

where $l$ is assumed to be a function of the joint positions and their derivatives and $\psi$ is the cost of the initial configuration. It is typically set to zero, unless the application requires to associate an explicit cost to it.

Assuming that the robot can reach the first drilling position in a safe configuration, far from collisions, the objective is to minimize the joint displacements, so as to avoid unnecessary risky movements [13]:

$$l(\mathbf{q}(i), \mathbf{q}(i-1)) = \sum_{k=0}^{n} |q_k(i) - q_k(i-1)| \quad (14)$$

where $k$ is the index spanning the joint position variables.

Now we can rewrite the objective function (13) in a recursive form and minimize over the possible configurations **q**, i.e.

$$I(0) = \psi(\mathbf{q}(0))$$
$$I_{\text{opt}}(i) = \min_{\mathbf{q}_{j,g} \in \mathcal{C}_{i-1}} [l(\mathbf{q}(i), \mathbf{q}(i-1)) + I(i-1)] \quad (15)$$

where $\mathcal{C}_{i-1}$ is the set of admissible $\mathbf{q}_{j,g}$ in $\mathcal{A}_{i-1}$ that also respect constraints on the derivative defined by the set $\mathcal{B}_i$. $I_{\text{opt}}(i)$ is termed *optimal return function*, which is the minimum value of the objective function if the process started at stage $i$, thus $I_{\text{opt}}(N_i)$ will be the optimized function.

## 2.3 Force ellipsoids

The dynamic programming setup described in Section 2.2 allows to compute an optimal solution for each sample of the redundancy parameter $u_j$ and inverse kinematic solution $g$ at the last stage $N_i$. The repetition of the recursion in (15) up until the last stage allows retrieving the globally-optimal solution for the given discretization of the domains. However, as discussed in Section 1, besides ensuring safe motions, we would like to guarantee stable drilling operations. Thus, among the safe trajectories that minimize the joint displacements, we choose the one that ends in the stiffest configuration. This selection is based on the analysis of the force ellipsoids.

The force ellipsoid represents the force transmission efficiency generated by the set of torque vectors with norm equal to one when the manipulator is in a given configuration [14]:

$$\boldsymbol{\tau}^T \boldsymbol{\tau} \leq 1 \Rightarrow \mathbf{f}^T \mathbf{J} \mathbf{J}^T \mathbf{f} \leq 1 \quad (16)$$

where $\boldsymbol{\tau}$ is the $n \times 1$ vector of actuation torques and **f** is the $m \times 1$ vector of forces and torques at the end-effector. The relationship $\boldsymbol{\tau} = \mathbf{J}^T \mathbf{f}$ has been used in the equation above, while the dependence of **J** on **q** has been omitted.

The shape of the ellipsoid is described by the eigenvectors **v** and the eigenvalues $\boldsymbol{\lambda}$ of the $m \times m$ matrix $\mathbf{A} = \mathbf{J} \mathbf{J}^T$. With reference to Figure 2, the eigenvectors indicate the orientation of the ellipsoid's axes with respect to the reference frame of the Jacobian and the reciprocals of the square roots of the eigenvalues represent the length of the semi-axes.

Assuming that the end-effector frame is aligned to the task frame in such a way that the $z$ axes are opposite and the $x$ axes are aligned, as in Figure 3, the objective is to maximize the force capacity along the $z$ axis, i.e. the drilling direction. As highlighted in [15], the best performances would be achieved if the ellipsoid's major axis and the direction of interest ($z$ in our case) had the same orientation. However, this is a difficult condition to obtain due to collisions and kinematic constraints. As a consequence, if **J** is computed with respect to the end-effector frame, we maximize the distance between the center of the frame, i.e. the center of the ellipsoid, and the intersection between the ellipsoid surface and the $z$ axis (the distance represented in yellow in Figure 2). This distance is given by the components of the $\mathbf{J} \mathbf{J}^T$ matrix selected through the vector $\boldsymbol{\eta}$, corresponding to the direction along which the stiffness has to be maximized.

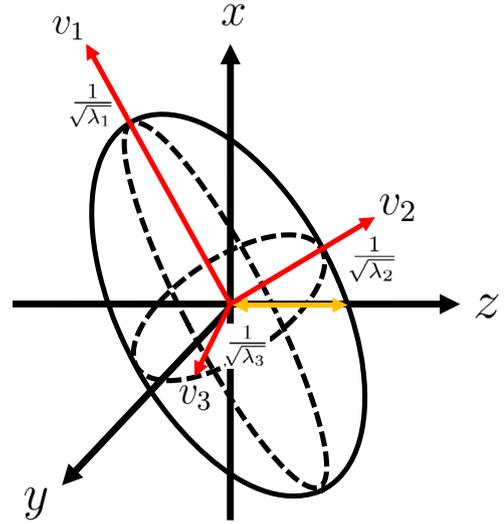

**Fig. 2.** Force ellipsoid representation.

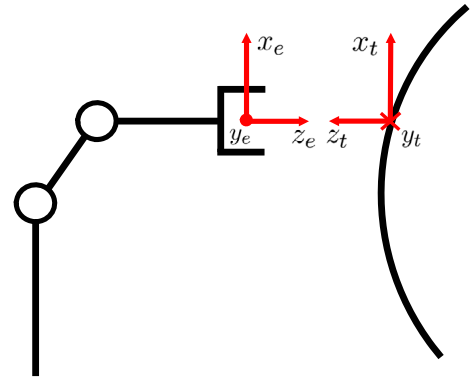

**Fig. 3.** End-effector and task frames.

All the valid joint configurations at the last waypoint are evaluated and the one with the maximum stiffness is retrieved by using the *Manipulator Mechanical Advantage* (MMA) index [16], i.e.

$$\mathbf{q}(N_i) = \arg \max_{\mathbf{q}_{j,g} \in \mathcal{C}_{N_i}} \left[ \left( \boldsymbol{\eta}^T \mathbf{J} \mathbf{J}^T \boldsymbol{\eta} \right)^{-1/2} \right] \quad (17)$$

This way, the only admissible configuration at the last stage is the stiffest one $\mathbf{q}(N_i)$. Since the Jacobian is expressed in the end-effector frame and we only consider the translational components, the $\boldsymbol{\eta}$ vector will be equal to $\boldsymbol{\eta} = [0 \ 0 \ 1 \ 0 \ 0 \ 0]^T$, which corresponds to the $z$ axis. It is worth remarking that (17) is a special case of the optimization problem in [17], obtained by setting the same compliance for all joints.

Recalling the optimization function (15) at the final stage $N_i$, we have that

$$I_{\text{opt}}(N_i) = \min_{\mathbf{q}_{j,g} \in \mathcal{C}_{N_i-1}} [l(\mathbf{q}(N_i), \mathbf{q}(N_i - 1)) + I(N_i - 1)] \quad (18)$$

where $\mathbf{q}(N_i)$ is the stiffest configuration obtained by applying (17).

## 2.4 Algorithmic implementation

The algorithm in [8,18] has been modified to include the stiffness optimization on the last waypoint, as well as to consider the specific objective function and constraints that characterize the LABOR use case. Its pseudo-code is provided in Algorithm 1.

The algorithm works on $N_g$ grids of size $N_i \times N_j$, whose cells $(i, j)$ represent possible configurations $\mathbf{q}$ at which the robot can be at the corresponding stage $i$. For each cell, a transition is evaluated towards all the cells at the next stage, i.e. $i+1$: constraints are checked and, if satisfied, the local cost $l$ and the cumulative cost $I$ are computed and saved. The optimal transition is the one returning the minimum cumulative cost.

Once all the costs have been computed for the last stage $N_i$, the stiffness of each enabled configuration at this stage is computed. The node with the maximal stiffness is selected and, starting from such configuration, the entire trajectory is built backwards, following the predecessors' list.

It is worth noticing that the grids represent the null space for the entire trajectory and their graphical representation through colored maps [8] is used for the a posteriori verification of the redundant parameter representativeness, as mentioned in Section 2.1.

## 3 Experimental results

Algorithm 1 has been implemented in ROS (Robot Operating System) in order to reuse the available modules to plan and visualize trajectories in the task space and to simulate the motion in the *RViz* virtual environment and assess the stiffness optimization. ROS has also been used to connect the DP planner to the *CoppeliaSim* simulator in order to command the Fanuc M20iA/35M model in the 3D scene with the LABOR panel.

The position and velocity limits that have been used in the algorithm are extracted from the official Fanuc datasheet and reported in Table 1 for the sake of clarity.

---

**Algorithm 1** Discrete DP redundancy resolution algorithm with stiffness optimization.

1: *Initialize* $\mathcal{A}_i, \forall i = 0..N_i$
2: *Initialize* $\mathcal{B}_i, \forall i = 1..N_i$
3: *Initialize* $\mathcal{C}_i = \oslash, \forall i = 0..N_i$
4: *Initialize costs* $I_{i,j,g} = +\infty, \forall i = 1..N_i, \forall j = 0..N_j, \forall g = 1..N_g$
5: *Initialize costs* $I_{0,j,g} = 0, \forall j = 0..N_j, \forall g = 1..N_g$
6: $\mathcal{C}_0 \leftarrow \mathcal{A}_0$
7: **for** $i \leftarrow 0$ **to** $N_i - 1$ **do**
8:     **for each** $\mathbf{q}_{j,g} \in \mathcal{C}_i$ **do**
9:         **for each** $\mathbf{q}_{k,h} \in \mathcal{A}_{i+1}$ **do**
10:            $\dot{\mathbf{q}} \leftarrow \frac{\mathbf{q}_{k,h} - \mathbf{q}_{j,g}}{\tau}$
11:            **if** $\dot{\mathbf{q}} \in \mathcal{B}_{i+1}$ **then**
12:                $\mathcal{C}_{i+1} \leftarrow \mathcal{C}_{i+1} + \{\mathbf{q}_{k,h}\}$
13:                *Compute local cost function* $l$
14:                **if** $I_{i,j,g} + l < I_{i+1,k,h}$ **then**
15:                    $I_{i+1,k,h} \leftarrow I_{i,j,g} + l$
16:                    *Set* $\mathbf{q}_{j,g}$ *at* $i$ *as predecessor of* $\mathbf{q}_{k,h}$ *at* $i+1$
17: $s_{max} \leftarrow -\infty$
18: **for each** $\mathbf{q}_{j,g} \in \mathcal{C}_{N_i}$ **do**
19:     *Compute the stiffness* $s_{j,g} \leftarrow (\boldsymbol{\eta}^T \mathbf{J}(\mathbf{q}_{j,g}) \mathbf{J}^T(\mathbf{q}_{j,g}) \boldsymbol{\eta})^{-1/2}$
20:     **if** $s_{j,g} > s_{max}$ **then**
21:         $s_{max} \leftarrow s_{j,g}$
22:         $\overline{\mathbf{q}} \leftarrow \mathbf{q}_{j,g}$
23: $I_{opt}(N_i) \leftarrow I_{N_i,j,g}$, *with* $j, g : \mathbf{q}_{j,g}(N_i) = \overline{\mathbf{q}}$
24: *Build the sequence of configurations* $\mathbf{q}(i)$ *starting from* $\mathbf{q}_{j,g}(N_i)$ *and going backwards in the predecessors list*

**Table 1.** Fanuc M20iA/35M joint limits.

|  | Joint 1 | Joint 2 | Joint 3 | Joint 4 | Joint 5 | Joint 6 | Slide joint |
|---|---|---|---|---|---|---|---|
| $q_{max}$ | −3.225 rad | −1.7553 rad | −3.225 rad | −3.49 rad | −2.4435 rad | −7.854 rad | −2.1 m |
| $q_{min}$ | 3.225 rad | 2.7925 rad | 4.8171 rad | 3.49 rad | 2.4435 rad | 7.854 rad | 2.1 m |
| $\dot{q}_{max}$ | 3.14 rad/s | 3.14 rad/s | 3.49 rad/s | 6.11 rad/s | 6.11 rad/s | 6.98 rad/s | 9.6 m/s |

**Table 2.** Time estimated for automatic operations on a CFRP and thermoplastic compound panel with 1500 holes using 9 mm grip Hi-lite fasteners.

| Assembly operations | Percentage of time | Per hole |
|---|---|---|
| Drilling: 3.1 mm diameter hole | 0.07 | 2.04 s |
| Diameter increasing from 3.1 to 4.0 mm | 0.06 | 1.93 s |
| Diameter increasing from 4.0 to 4.8 mm | 0.06 | 1.93 s |
| Countersinking | 0.12 | 3.54 s |
| Sealing and riveting | 0.55 | 16.38 s |
| Inspection | 0.14 | 4.19 s |
| Total time | 1.00 | 30 s |

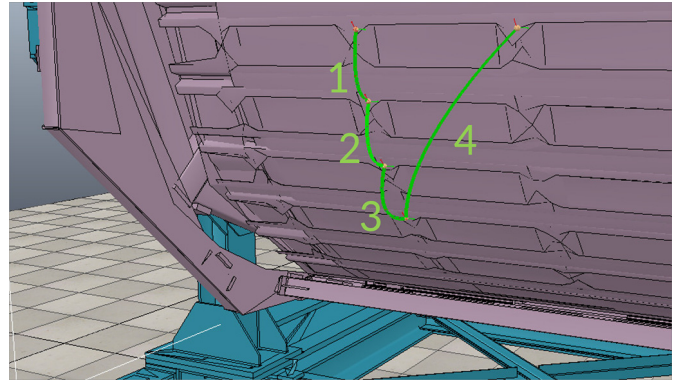

**Fig. 4.** Four task-space trajectories in the CoppeliaSim simulation environment.

## 3.1 Time requirement

The dynamic programming algorithm presented in Section 2.2 is usually employed for off-line applications [8,13,19,20] because of the computational effort required to find the optimal solution. While the algorithm is far from being applied in scenarios requiring real-time planning, in some circumstances, it is suitable to plan trajectories while the robot already moves or interacts with the workpiece. In the LABOR project, the trajectories cannot be completely pre-planned because they are subject to the referencing operations performed by the internal robot, that provides the initial drilling position of a known pattern. However, at each position in the pattern, the robot stops to perform the task of interest (drilling/sealing/fastening) and this time can be used to plan the next hole-to-hole trajectory in the pattern in an optimal way: the trajectory optimization process must not exceed the time for drilling and sealing/fastening operations. Here we analyze what, in the LABOR project, these time requirements are, thus providing an upper bound for the DP algorithm to complete before the external robot needs to move to the next hole.

The estimation is based on the requirement that the cycle time is 30 s per hole (considering drilling, countersinking, hole inspection, sealing, fastener insertion and not considering fastener inspection), but such a time is not allocated to specific tasks. Thus, we make the simplifying assumption that the distribution of time across the several tasks is the same as manual operations. We consider the time needed for the manual installation of 9 mm grip Hi-lite fasteners for 1500 holes in a CFRP and thermoplastic compound panel, and keep the same time percentages for the automatic assembly. The results are shown in Table 2.

In particular, the drilling time results to be equal to 13.62 s (considering all the operations except sealing and riveting) and the fastening time is equal to 16.38 s (considering only sealing and riveting). As a consequence, the planning for each hole-to-hole trajectory has to be completed within the minimum of such times.

## 3.2 Planned trajectories

Simulations have been performed by planning four hole-to-hole trajectories for five drilling points, by keeping the orientation fixed. Such trajectories are longer than the real ones to highlight some characteristics of the optimization process that we will discuss next. The trajectories are shown in Figure 4, and the parameters used for planning are reported in Table 3. The initial position for the first trajectory is not known so it is computed as a result of the optimization process, while, for the others, the initial position must correspond to the final position of the previous trajectory.

The slide position has been empirically selected as redundancy parameter and, as discussed in Section 2.1, an a posteriori verification is performed to be sure that the condition (5) holds for all waypoints. The *slide resolution* in Table 3 indicates the minimum displacement of the slide joint in its discrete domain and has been selected as a result of the trade-off between optimization quality and execution time: the higher the resolution is, the closer the solution will be to the global optimum. Like many other six-axis industrial robots, even for the Fanuc M20iA/35M, it is $N_g = 8$ although, due to constraints, the number of admissible inverse kinematics solutions may be lower.

The execution of the four trajectories of Figure 4 in CoppeliaSim is shown in [21]: unlike the current planning policy, the DP algorithm makes the slide move to minimize the overall motion of the joints and to reach a stiff configuration at the end of each trajectory. It is important to remark that this is done by assigning the tasks directly in the task space. Joint limits and collisions are automatically checked and avoided. The planning time spans from 21 to 29 s for trajectories longer than 288 mm with 10/15 waypoints. The trajectories that are considered in the LABOR project are much shorter (25.4 mm), being the points in the pattern much closer. By keeping the same trajectory resolution, we would only need 3 waypoints to describe such trajectories. Since the computation time scales linearly with the number of waypoints, we should expect the planning time to be equal to about 8 s, respecting the upper bound of 13.62 s as estimated above. Plots of planned joint space trajectories against time are provided in Figure 5.

**Table 3.** DP solver parameters for trajectories 1–4. Planning times are related to the execution of the algorithm on a virtual machine with a single-threaded implementation.

| | Trajectory | Waypoints ($N_i$) | Length (mm) | Slide resolution (mm) | Duration (T) (s) | Planning time (s) |
|---|---|---|---|---|---|---|
| 1 | 1st to 2nd hole | 10 | 288.5 | 13.2 | 0.55 | 27 |
| 2 | 2nd to 3rd hole | 10 | 321.4 | 13.2 | 0.55 | 21 |
| 3 | 3rd to 4th hole | 10 | 339.0 | 13.2 | 0.55 | 29 |
| 4 | 4th to 1st lateral hole | 15 | 1060.0 | 13.2 | 0.55 | 24 |

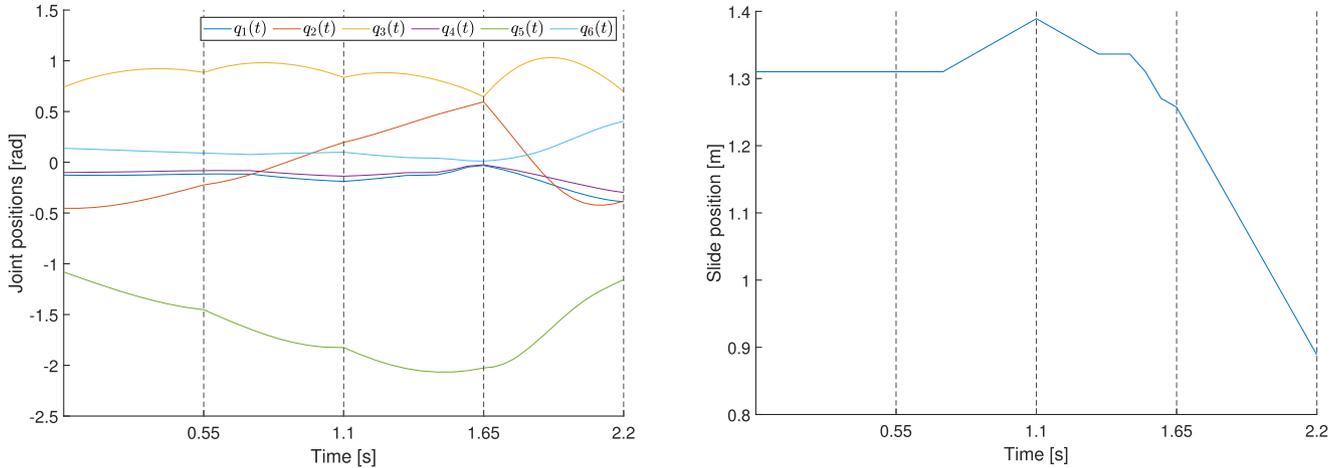

**Fig. 5.** Planned joint and slide positions for trajectories 1–4 plotted in sequence. The end of each trajectory is indicated with a dotted line.

### 3.3 Stiffness analysis

Now let us make a deeper analysis on the stiffness considering the 4th trajectory. In Figure 6 (left), the force ellipsoid for the last position (the working pose) is plotted in RViz. Here the red, green and blue vectors are the ellipsoid's semi-axes, while the yellow vector represents the distance between the center of the tool frame and the ellipsoid surface, to be maximized. The stiffest configuration is the one having the yellow vector aligned with the major axis of the ellipsoid, but kinematic constraints and collisions may prevent to reach such a condition. For the 4th trajectory, the resulting MMA is equal to 1.099. Details on the direction and length of each axis with respect to the tool frame (as the one represented in Figure 3) are reported in Table 4. It is worth noticing that the robot is slightly placed on the left of the working point, while the ellipsoid is quite oblique. In fact the $y$ component of the major semi-axis (the red one) is equal to $-0.0825$ while its $z$ component is almost aligned with the $z$ axis of the tool frame, i.e. $-0.9008$. Hence, it is intuitive to conclude that stiffer configurations may exist.

Let us repeat the experiment by considering a slower end-effector trajectory (from 0.55 to 1.4 s), so that the robot will have more time to move its kinematic structure without violating joint velocity and acceleration limits. The final configuration is shown in Figure 6 (right), together with the force ellipsoid. In this case the MMA is equal to 1.127, which is higher than the MMA obtained with a faster trajectory. The related ellipsoid's parameters are reported in Table 5. We can notice that the ellipsoid is more elongated and its major semi-axis is better aligned with the drilling direction, resulting in a stiffer posture. The $y$ component is perfectly aligned with the corresponding axis of the tool frame and its length is also higher (1.8 instead of 1.7629).

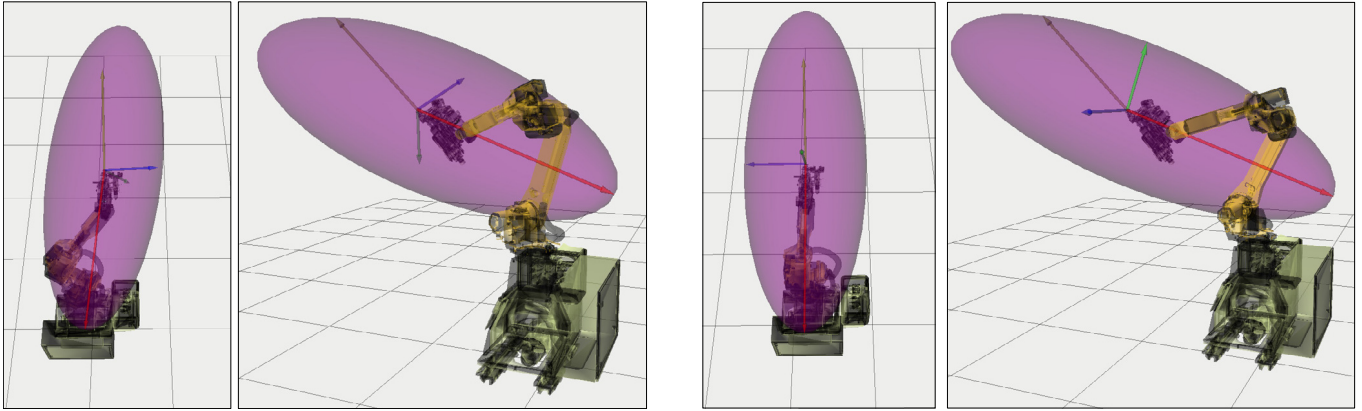

**Fig. 6.** Frontal and lateral view of the force ellipsoid of the last position of the 4th trajectory, planned with a trajectory duration of 0.55 s (left) and 1.4 s (right).

**Table 4.** Parameters of the ellipsoid resulting from the planning of the 4th trajectory with duration 0.55 s, expressed with respect to the tool frame.

| Axis | Eigenvector | | | Length |
|---|---|---|---|---|
| | x | y | z | $(1/\sqrt{\lambda})$ |
| **Red axis** | 0.4262 | −0.0825 | −0.9008 | 1.7629 |
| **Green axis** | −0.8127 | 0.4025 | −0.4214 | 0.5726 |
| **Blue axis** | 0.3973 | 0.9117 | 0.1045 | 0.6608 |

**Table 5.** Parameters of the ellipsoid resulting from the planning of the 4th trajectory with duration 1.4 s, expressed with respect to the tool frame.

| Axis | Eigenvector | | | Length |
|---|---|---|---|---|
| | x | y | z | $(1/\sqrt{\lambda})$ |
| **Red axis** | 0.4193 | 0.001 | −0.9079 | 1.8 |
| **Green axis** | 0.9077 | −0.0183 | 0.419 | 0.5743 |
| **Blue axis** | −0.0161 | −0.9998 | −0.008 | 0.6613 |

A comparison between the two trajectories is provided in [22]: in the second trajectory the robot is obviously slower but this allows reaching a stiffer final posture.

Hence, depending on the assigned parameters and the trajectory characteristics, the joint-space trajectory will change, allowing the robot to adapt to different situations and scenarios.

### 3.4 Comparative results

As explained in Section 1.1, the traditional approach is based on the heuristics of fixing the slide in a position which should lead to the stiffest manipulator configuration. According to it, in the case of the first three trajectories of Table 3, the slide is placed in front of the vertical hole sequence and the joint space trajectory is found. We remark that no optimization occurs in this case, as the inverse kinematics problem is squared. In terms of both MMA and joint displacements, the two techniques provide similar results, as Figure 7 and Table 6 testify.

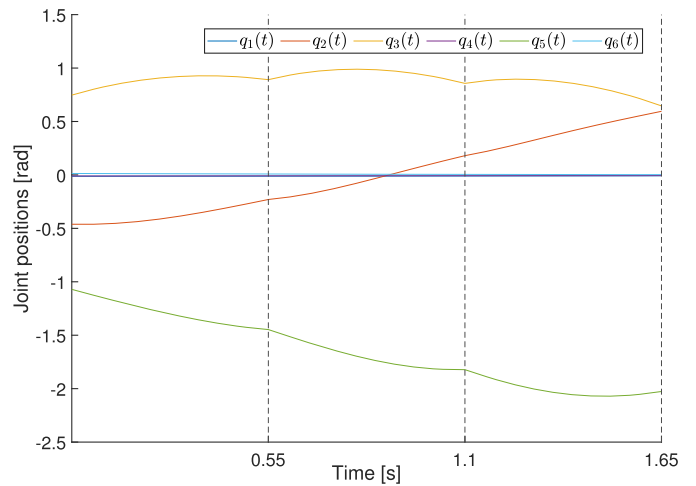

**Fig. 7.** Planned joint positions with fixed slide for the trajectories 1–3 plotted in sequence. The end of each trajectory is indicated with a dotted line.

**Table 6.** MMA values obtained from the dynamic programming algorithm moving the slide and from the heuristics by fixing the slide position.

| Trajectory | Moving slide | Fixed slide |
|---|---|---|
| 1 | 0.922 | 0.922 |
| 2 | 0.819 | 0.819 |
| 3 | 0.741 | 0.741 |

Now, let us consider the 4th trajectory again, which is a lateral motion. This task cannot be achieved by fixing the slide, as the path exits the manipulator's workspace. In the current LABOR setup, as shown in [23], the robot returns to its home position, the slide moves in front of the lateral hole, then the robot arm moves to the drilling position. Alternatively, inverse kinematics could be solved for the last hole, assuming the slide already be in front of it: two point-to-point motions are planned for both the slide and the arm, commanded separately, and executed at the same time. Regardless of the time difference of the two solutions, either strategy does not guarantee any control over the end-effector motion, resulting in a safety issue. On the contrary, in the proposed DP technique, the path is defined in the workspace and the movements of both the arm and the slide are planned together: a collision-free safe motion of both the end-effector and the whole kinematic chain is guaranteed, together with a final posture that is at least as stiff as the one obtained with the traditional approach.

In general, it should be expected that the benefits of the proposed techniques are much more evident when the assembly operations require lateral motions along the fuselage panel or coverage or more complex geometries. For instance, let us consider the assembly operations around a window of the fuselage. In [23], we show what the movement of the arm and the slide would be if the trajectories from one hole to the other, along the window perimeter, were planned with the proposed DP algorithm. The slide would follow the manipulator for each hole to be drilled so as to minimize the global joint motion and allows for less elongated postures of the arm when reaching lateral holes. On the contrary, by fixing the slide position, the manipulator must elongate to reach all the holes, to the detriment of stiffness. In addition, due to the higher joint displacements, the time required to execute the path is about three times higher than dynamic programming.

## 4 Conclusions

Robotized solutions in large industrial plants are typically characterized by the presence of slides on which robots are mounted to increase their workspace. This leads to the introduction of kinematic redundancy which, however, is not efficiently handled and not exploited to satisfy typical requirements of aerospace manufacturing that are very demanding and sometimes not achievable by the common industrial robots. An example of such solutions is given by the LABOR project which provides a system to make autonomous assembly operations of fuselage panels by using two cooperating robots placed on slides. Since strict safety, stability, accuracy and efficiency requirements exist, we proposed a methodology that retrieves optimized trajectories by exploiting the kinematic redundancy provided by the system itself. Joint space trajectories that satisfy the requirements above are generated in the allotted planning time. In particular, we planned safe motions for the robots by minimizing the joint displacements, and maximizing the stiffness at the working pose. At the same time, we avoided collisions, self-collisions and joint limits. We employed a discrete dynamic programming algorithm, characterized by a high degree of flexibility and adaptability to different scenarios. This is the main strength of the proposed technique as it allows to tackle different situations with different robots, trajectories and panels, by only changing, for example, the cost function and the constraints. Moreover, a suitable parametrization of the algorithm allows to achieve the desired performance and comply with the time requirements.

We have seen that the LABOR cell is provided with two robots working together to perform assembly operations. We expect that the workcell efficiency can be further improved if trajectories were jointly planned for the two robots at the same time. It is, therefore, our plan to extend the framework to cooperating scenarios, which will be the subject of future research.

*Acknowledgments.* This work has been funded by the European Commission under the CleanSky 2 project LABOR (GA n. 785419), with Leonardo S.p.A. as Topic Manager.